\documentclass[10pt,twocolumn,letterpaper]{article}

\usepackage{cvpr}
\usepackage{times}
\usepackage{epsfig}
\usepackage{graphicx}
\usepackage{amsmath}
\usepackage{amssymb}

\usepackage[breaklinks=true,bookmarks=false]{hyperref}

\cvprfinalcopy 

\def\cvprPaperID{****} 

\begin{document}
	
	\title{Learning Context Graph for Person Search}
	
	\newcommand*\samethanks[1][\value{footnote}]{\footnotemark[#1]}
	
	
	
	
	\author{Yichao Yan$^{1,2,3,4}$\thanks{Yichao Yan and Qiang Zhang contribute equally to this paper} \qquad 
		Qiang Zhang$^{1}$\footnotemark[1] \qquad
		Bingbing Ni$^{1}$\thanks{the corresponding author is Bingbing Ni} \qquad \\
		Wendong Zhang$^{2}$ \qquad 
		Minghao Xu$^{1}$ \qquad 
		Xiaokang Yang$^{2}$ \qquad 
		\\
		\noindent
		$^{1}$Shanghai Jiao Tong University, China \qquad \\
		$^{2}$MoE Key Lab of Artificial Intelligence, AI
		Institute, Shanghai Jiao Tong University, China \qquad \\
		$^{3}$ Tencent YouTu Lab, China \qquad 
		$^{4}$ Inception Institute of Artificial Intelligence, UAE \\
		$^{}$ {\tt\small \{yanyichao, zhangqiang2016, nibingbing, diergent, xuminghao118, xkyang\}@sjtu.edu.cn} \\
	}

	\maketitle
	
	\begin{abstract}
		Person re-identification has achieved great progress with deep convolutional neural networks. However, most previous methods focus on learning individual appearance feature embedding, and it is hard for the models to handle difficult situations with different illumination, large pose variance and occlusion. In this work, we take a step further and consider employing context information for person search. For a probe-gallery pair, we first propose a contextual instance expansion module, which employs a relative attention module to search and filter useful context information in the scene. We also build a graph learning framework to effectively employ context pairs to update target similarity. These two modules are built on top of a joint detection and instance feature learning framework, which improves the discriminativeness of the learned features. The proposed framework achieves state-of-the-art performance on two widely used person search datasets.  
	\end{abstract}
	
	\section{Introduction}
	
	Persons re-identification (re-id) is a fundamental and important research topic in computer vision. It aims to re-identify individuals across multi-camera surveillance systems. Person re-identification has great potential in applications related with video surveillance, such as searching for lost people or suspects. These applications are closely related to public security and safety, therefore person re-id has been receiving increasing attentions over recent years. 
	For a typical person re-id pipeline, the re-id system is provided with a target person as probe and aims to search through a gallery of known ID recordings to find the matched ones. 
	Person re-id is extremely challenging due to the following reasons. First, distributions of probe and gallery are multi-modal due to different data sources. For example, the pedestrians can be captured by surveillance cameras or smart phones. Second, different illuminations and human poses will increase intra-class variations. Third, inaccurate detection/tracking, occlusions and background clutters lead to heavy appearance changes, which further increases the difficulty for person re-id. 
	
	\begin{figure}[t]
		\centering
		\includegraphics[width=\linewidth]{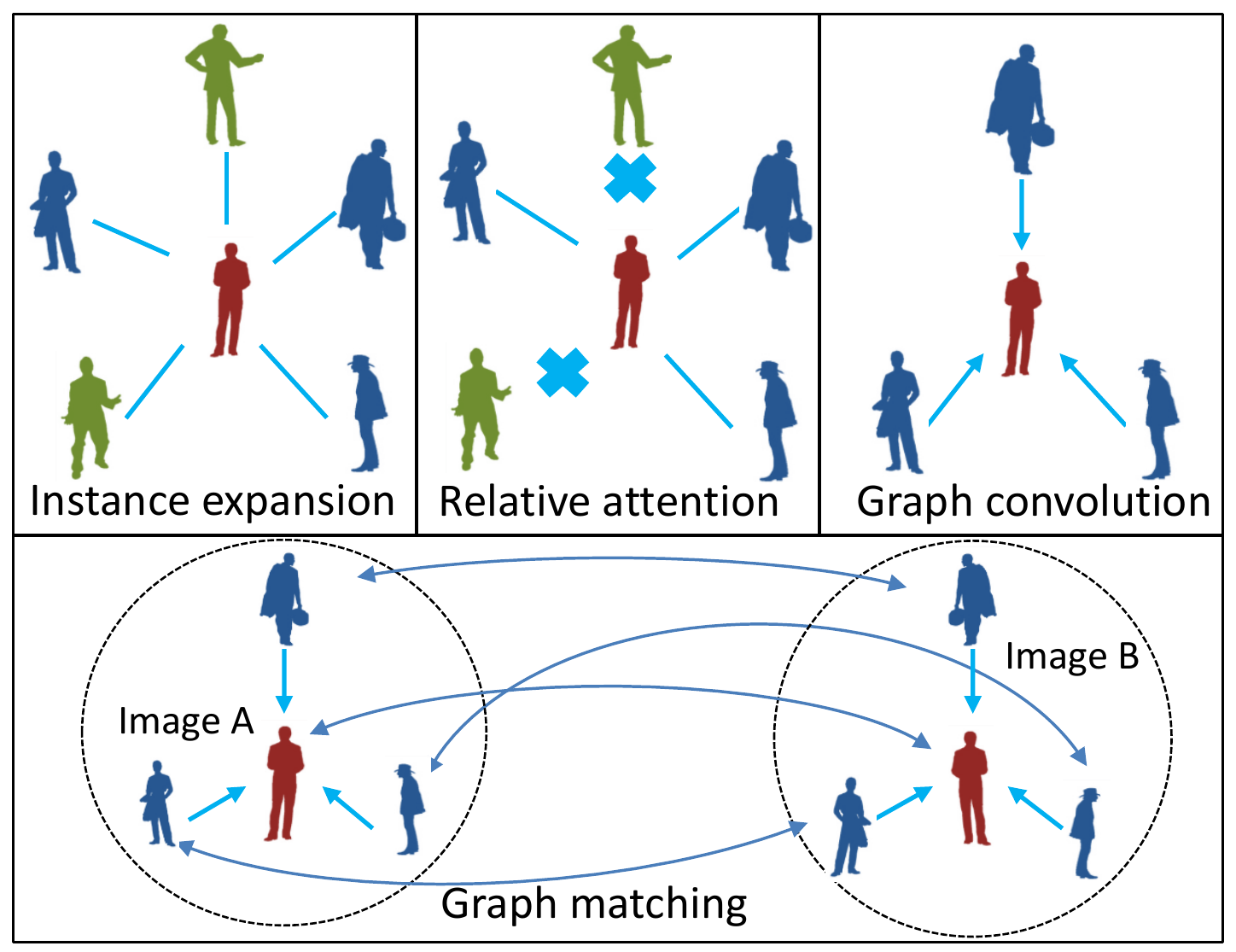}
		\caption{Illustration of the proposed framework.}
		\label{fig:intro}
	\end{figure}
	
	Traditional person re-id tasks only focus on matching manually cropped image snapshots or video clips across different cameras. These methods consider learning distance metrics based on individual features. Therefore, one important prerequisite is that the foreground pedestrian should be precisely detected or annotated in the scene. Otherwise, inaccurate detection or annotation will bring heavy noise to individual appearances, which makes this problem setting impractical in real scenarios. To bridge this gap, some recent works~\cite{DBLP:conf/mm/XuMHL14,DBLP:conf/cvpr/XiaoLWLW17} introduce the person search setting into this domain. The idea is to simultaneously handle two tasks (\ie, pedestrian detection and person re-identification) within a single framework. This setting is closer to real-world applications and allows the system to function without off-line pedestrian detectors. However, these methods still employ individual features as appearance cues. Hence it is hard to distinguish people with similar wearings, especially in situations where we have to search through a huge gallery set. To further address this issue, some recent works observe that scene context can provide significant richer information than individual appearance cues. In real situations, people are likely to walk in groups~\cite{DBLP:conf/avss/MazzonPC13}. Even when people are walking alone, other neighboring pedestrians appearing in the same scene also contain important context cues. In other words, co-travelers captured by the same camera will have high chance to be also captured by its neighboring cameras. Employing context/group information is a promising direction to tackle real-world person re-identification, however recent works suffer from the following issues. First, how to identify a group is not a trivial task. Existing methods~\cite{DBLP:conf/iccv/LisantiMBF17} typically utilize manual annotations to locate semantic groups, which requires extensive human labor. Other methods~\cite{DBLP:conf/iccvw/CaoCHP17,DBLP:conf/eccv/AssariIS16} make use of spatial and temporal cues such as velocity and relative position in the scene, which are considered as social constraints to model group behaviors to help facilitate person re-id. These social force models utilize elaborately designed constraints to simulate social influences in the scene, which usually does not have trivial solutions and is hard to optimize.  
	
	In this work, we propose a novel framework to explore context cues for robust person search.
	The overall pipeline of the proposed framework is illustrated in Figure~\ref{fig:intro}.
	As individual appearance features are not powerful enough to distinguish different people, we first propose to expand instance-level features with contextual information. 
	For person search, the most important contexts are neighboring co-travelers. Therefore, given a target person (marked red in Figure~\ref{fig:intro}), we collect all other pedestrians in the scene as context candidates. 
	Among all these candidates, some of the contexts are useful and others are just noises. Therefore, before utilizing context information, one important step is to filter useful contexts from noise ones. To this end, we introduce a relative attention module which takes in context candidates in both probe and gallery images, and outputs the matched pairs as informative context.
	With individuals and the corresponding contexts, the remaining question is how to make full use of all the information to make a more confident judgment whether the target pair belongs to the same identity. We propose to build a context graph to model the global similarity of probe-gallery pairs. 
	Specifically, graph nodes consist of a target pair and context pairs. To employ context information, all the context nodes are connected to the target node. This graph is trained to output the similarity of the target pair.
	
	We evaluate our framework on two widely used benchmarks, including CUHK-SYSU~\cite{DBLP:conf/cvpr/XiaoLWLW17} and PRW dataset~\cite{DBLP:conf/cvpr/ZhengZSCYT17}. Experiment results demonstrate that our method can achieve significant improvements over previous state-of-the-arts. Our contributions include: 
	1) We introduce a multi-part learning scheme into person search, which supports end-to-end human detection and multi-part feature learning.
	2) We introduce a relative attention model to adaptively select informative context the scene.
	3) We build a graph to learn global similarity between two individuals considering context information. 
	
	\section{Related Work}
	
	Person re-id aims to associate pedestrians over non-overlapping cameras. Most previous methods try to address this task on two directions, \ie, feature representation and distance metric learning. 
	Before deep learning methods gets popular, previous methods design different kinds of hand-crafted features, such as color~\cite{DBLP:journals/ijcv/Lowe04}, texture~\cite{DBLP:conf/cvpr/FarenzenaBPMC10} and gradient~\cite{DBLP:conf/eccv/GrayT08,BEDAGKARGALA20121908}. These methods achieve certain success on small datasets. However, the representation capability of hand-crafted features is limited for large-scale searching.
	Similar limitations also apply for traditional distance metric learning methods~\cite{DBLP:conf/cvpr/ZhangXG16,DBLP:conf/cvpr/KostingerHWRB12,CHENG2018}, which aim to optimize a distance function based on certain feature. However, the learned distance metric usually overfits on the training data. Therefore, the generalization ability of these methods is also limited. 
	With the renaissance of deep learning~\cite{DBLP:conf/nips/KrizhevskySH12,DBLP:conf/eccv/LuMNYRY18,DBLP:journals/tip/WangSS18,DBLP:conf/ijcai/YanNY17,DBLP:conf/mm/YanXNZY17}, CNN was first introduced to address person re-id in ~\cite{DBLP:conf/cvpr/ZhaoOW14,DBLP:conf/cvpr/LiZXW14}, and it quickly dominates this domain ever since. In recent years, a large amount of works have proposed different model structures~\cite{DBLP:conf/cvpr/ZhaoOW14,DBLP:conf/cvpr/LiuNYZCH18,DBLP:conf/cvpr/LiaoHZL15,DBLP:conf/eccv/YanNSMYY16,DBLP:conf/eccv/YanNSMYY16,DBLP:conf/cvpr/ChenYCZ16,DBLP:conf/cvpr/ChenCZH17,DBLP:conf/mm/SongNYRXY17,DBLP:conf/cvpr/ZhouHWWT17,DBLP:journals/pami/ZhengGX16}. Some methods enable end-to-end feature and metric learning~\cite{DBLP:conf/cvpr/AhmedJM15,DBLP:conf/cvpr/SchroffKP15,DBLP:conf/cvpr/ChenCZH17,ZHAO2018,DBLP:journals/corr/ZhengZY16}, other methods employ part information to build more robust representation~\cite{DBLP:conf/iccv/ShenLYXWW15,DBLP:journals/tip/LinSYXWWL17,DBLP:journals/tip/LinSYXWWL17,DBLP:journals/tip/LinSYXWWL17,DBLP:conf/iccv/SuLZX0T17,DBLP:conf/mm/WeiZY0T17,DBLP:journals/corr/ZhengHLY17,yan2018multi}. These approaches have achieved promising result on recent person re-id benchmarks. However, all these methods only focus on learning appearance features based on given human bounding box. In real applications, these methods should be combined with an offline pedestrian detector.
	
	To facilitate real-world person re-id, recent methods propose to jointly address the task of detection and re-identification~\cite{DBLP:conf/mm/XuMHL14,DBLP:conf/cvpr/XiaoLWLW17}. State-of-the-art methods~\cite{DBLP:journals/corr/abs-1804-00376,DBLP:conf/iccv/LiuFJKZQJY17,DBLP:journals/corr/XiaoXTHWF17} design online learning object functions to learn large number of identities in the training set. These methods achieve great performance on recent person search datasets. However, these methods only employ individual appearance for verification, which ignores the underlying relationship between individuals in the scene. Employing group/social information could be a promising direction to further improve system's performance. Although some methods~\cite{DBLP:conf/iccv/LisantiMBF17,DBLP:conf/iccvw/CaoCHP17,DBLP:conf/eccv/AssariIS16} have made efforts towards this direction for person re-id task, we design a novel context learning framework with graph model in person search scenario.
	
	Graph convolutional networks (GCN)~\cite{DBLP:journals/corr/KipfW16} has been proposed to learn graph relations with convolution, which facilitates the optimization of traditional graph model. GCN has been applied to various tasks~\cite{DBLP:conf/miccai/KtenaPFRLGR17,DBLP:journals/corr/abs-1802-07459,DBLP:conf/eccv/QiWJSZ18,DBLP:conf/eccv/WangG18,DBLP:journals/corr/abs-1711-04043}. 
	Several recent works~\cite{DBLP:conf/cvpr/Chen0LSW18,DBLP:conf/eccv/ShenLYCW18} employ graph model for person re-id, but the graphs are constructed to model relationships between probe-gallery pairs, and no context information is considered. In this work, we design a target-context graph and employ a pairwise GCN to learn visual relations in the scene.

	\begin{figure*}[t]
		\centering
		\includegraphics[width=\linewidth]{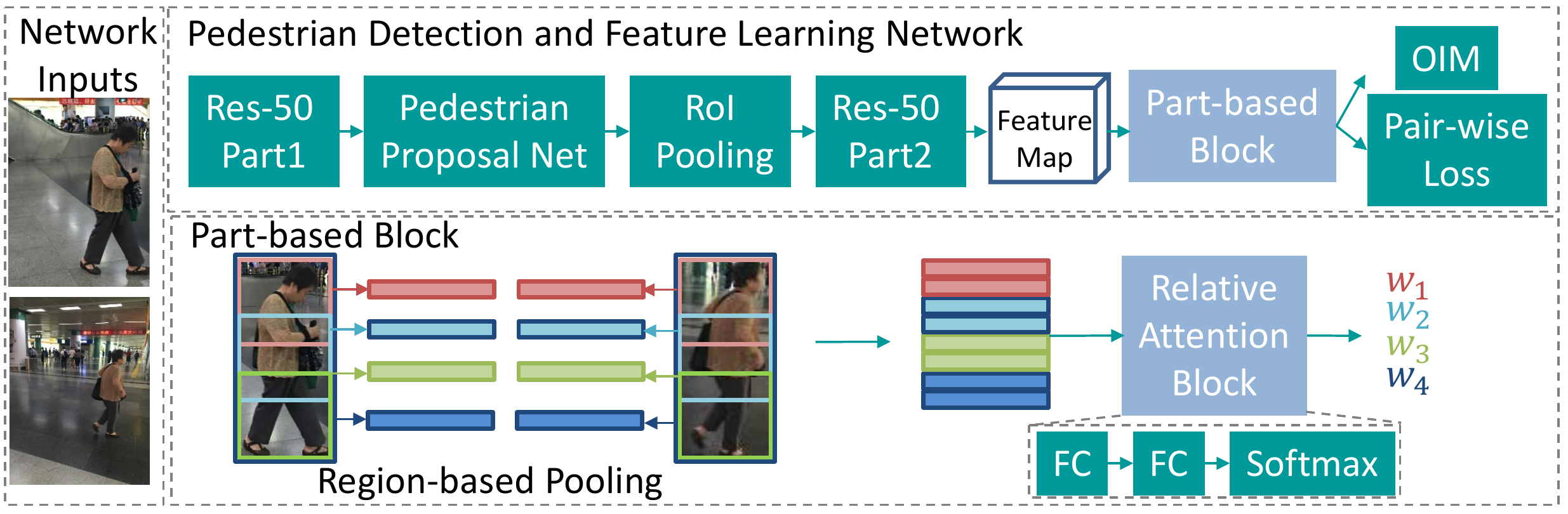}
		\caption{Architecture of the detection and part-based feature learning framework.}
		\label{fig:model1}
	\end{figure*}
	
	\section{Methodology}
	
	\subsection{Overview}
	Although deep CNN models have greatly improved the representation ability of instance-level individual features, it is still difficult to retrieve the target persons across different camera views in many complex situations. Therefore, our core idea is to expand instance features such that context information can be used to learn better representation.
	Specifically, our framework is consists of three major steps.
	
	\textbf{Instance Detection and Feature Learning}. In this stage, we utilize a baseline CNN to perform joint detection and feature learning on person search datasets. Following Faster R-CNN~\cite{DBLP:conf/nips/RenHGS15} framework, a region proposal network (RPN) is embedded on top of feature maps generated by ResNet-50 baseline. The bounding boxes are then fed into an RoI-Pooling layer to extract individual appearance representation. In addition, we introduce part-based feature learning framework into our model, thus yielding more discriminative representation. 
	
	\textbf{Contextual Instance Expansion}. This is one of the key components of our framework, which is built to expand instance feature with context information for better representation. All the instance pairs between query and gallery images are considered as context candidates, and noise contexts needs to be filtered. To this end, we build a relative attention layer to measure the visual similarity between context pairs and only the pairs with sufficient confidence are picked out as informative contexts.
	
	\textbf{Contextual Graph Representation Learning}. This is another crucial component of our framework. Given a probe-gallery pair, we construct a graph to measure the similarity of target pair. Graph nodes consist of target persons and the associated context pairs, which are connected with graph edges. We apply a graph convolutional network to learn the similarity between probe-gallery pair.

	\subsection{Instance Detection and  Feature Learning}
	\subsubsection{Pedestrian Detection}
	
	Real person search scenarios are usually in the wild, therefore target pedestrians need to be detected in the scene before searching can be performed. Recent state-of-the-art frameworks perform person detection and feature learning in a single framework, which significantly facilitates traditional pipelines (\ie, separate detection and feature learning). In this work, we take this popular structure as backbone network in our framework. The overall detection and feature learning framework is illustrated in Figure~\ref{fig:model1}.
	
	Specifically, we employ ResNet-50~\cite{DBLP:conf/cvpr/HeZRS16} as stem, which is divided into two parts. The first part (conv1 to conv4\_3) outputs 1024 channel feature maps, which have 1/16 resolutions of the input image. Following Faster R-CNN framework, a pedestrian proposal network(PPN) is built on top of these feature maps to generate person proposals, which is further passed to a $512\times3\times3$ convolutional layer to generate pedestrian feature representation. Similar to previous frameworks, we assign 9 anchors to each feature map. Two loss terms are utilized to train the PPN, \ie, a binary Softmax classifier to judge whether the anchor is a person or not and a linear layer to perform bounding box regressions. Finally, non-maximum suppression is used to remove duplicated detections and 128 proposals are kept for each image.  
	All the candidate proposals are fed into a RoI-Pooling layer to generate feature representations for each bounding box. These features are then convolved by the second part of ResNet50 (conv4\_4 to conv5\_3). These features are then connected to an average pooling layer to generate 2048-dimensional feature representation. The pooled features are connected with two fully connected (Fc) layers. The first branch is a binary Softmax layer which is trained to make person/non-person judgments. The second branch is a 256-dimensional Fc layer, whose outputs are further L-2 normalized as feature representation for inference. 
	
	\subsubsection{Region-based Feature Learning}
	
	Part-based models have proven to be effective for person re-id tasks, which motivates us to also consider modeling human parts for person search task. Therefore, we design a region based learning framework to effectively model part features. In addition to a global average pooling layer, we design several part-based pooling layers after the second part of ResNet-50. Each part-based pooling layer concentrates on a specific human part and pools the features into a 2048-dimensional vector, which are further connected with a fully connected layer and normalized into 256-dimensional feature. These features are further used for feature learning. Specifically, we design 3 part-sensitive pooling layers, which focus on upper-body, torso and lower-body. Each layer pools a 7$\times$3 region in the 7$\times$7 feature maps, which is illustrated in Figure~\ref{fig:model1}.
	
	To learn robust feature representation, the designed loss terms should guarantee the discriminativeness of the learned feature. 
	Although Softmax loss is wildly employed for classification tasks, it is hard to train a Softmax layer when the identity number is large. 
	Meanwhile, faster R-CNN framework consumes large memory, which limits that the mini-batch can only have small size. 
	Hence, the identities appeared in each mini-batch are highly sparse, which makes Softmax loss training even harder.
	To address this issue, previous methods design several online learning loss. In this work, we adopt online instance matching (OIM) loss~\cite{DBLP:conf/cvpr/XiaoLWLW17} to supervise feature learning for each part.
	
	\begin{figure*}[t]
		\centering
		\includegraphics[width=\linewidth]{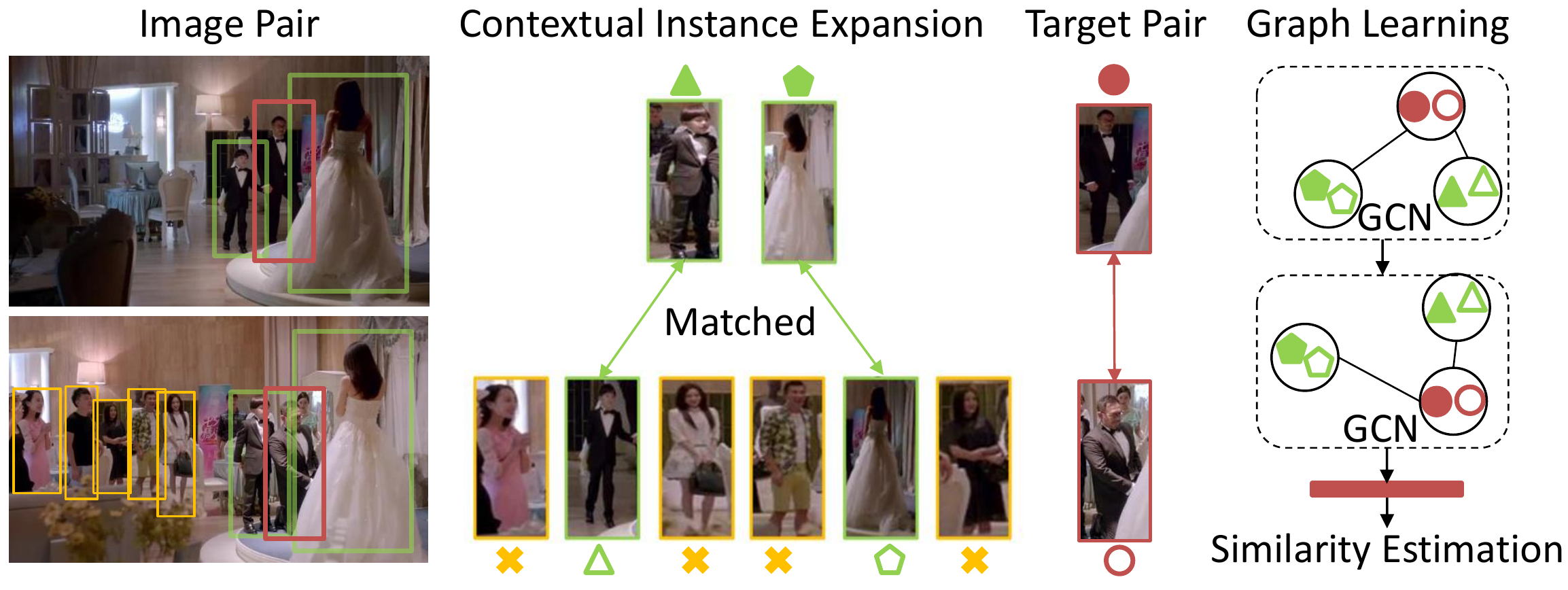}
		\caption{Architecture of the detection and part-based feature learning framework.}
		\label{fig:model2}
	\end{figure*}
	
	\subsection{Contextual Instance Expansion}
	As individual features are not sufficient for real world person retrieval task, we propose to employ context information as complement.  
	An example is illustrated in Figure~\ref{fig:model2}. 
	The objective is to identify whether the men in red bounding boxes belong to a same identity. However, the results are usually not confident as the appearance of the person suffer from great variation across different scenes. In this case, we observe that the same persons in green bounding boxes appear in both scenes, thus a more confident judgment can be made that the men in red bounding boxes do belong to the same identity. Therefore, the persons in green bounding boxes play a positive role, while other persons in the scene are noise contexts. In this part, we propose a relative attention model to filter all the contexts and only positive contexts are selected to expand individual features.
	
	Specifically, we consider the set of persons which appear on both probe and gallery scenes as positive contexts. 
	The remaining question is how to judge whether two detected pedestrians belong to the same identity. A trivial solution is to compute the similarity between the feature pairs, and set a threshold to make binary decision. 
	We use $\mathbf{x}_i^r$,$\mathbf{x}_j^r$ to denote the features of the $r-$th part from object $i$ and $j$.
	Consider different object parts, the overall similarity $s(i,j)$ can be represented as the summation of different parts:
	\begin{equation}\label{equ_relative}
		s(i,j) = \sum_{r=1}^{R} w_r cos(\mathbf{x}_i^r,\mathbf{x}_j^r),
	\end{equation} 
	where R is number of part (R=4 in our framework), and $cos(\mathbf{x}_i,\mathbf{x}_j)$ denotes the cosine similarity between feature pairs. $w_r$ is the contribution of the $r$-th object part, which is usually set by empirical experiences.  
	However, as discussed in~\cite{DBLP:journals/corr/abs-1806-03084}, uniformly combining these terms is not the optimal solution. Because the contributions of different object parts are significantly different across samples, due to possible occlusions, different viewpoints and lighting conditions. Hence, Huang \etal~\cite{DBLP:journals/corr/abs-1806-03084} propose an instance region attention network to assign different weights to instance parts. The attention weights measure the instance-wise part contributions, and part similarity is multiplied by both parts' attention weights. 
	In this work, we observe that part contributions are not only related to sample part itself, but are also related to the part to be matched. In other words, part contribution is related to part pairs. An example is illustrated in Figure~\ref{fig:model1}. 
	Head and facial appearance are important cues for the first body part. However, when frontal view is not available, both parts tend to have low attention weights. In fact, these two parts share great similarity and can provide important information to make a positive judgment. Motivated by this observation, we design a relative attention network which considers pair-wise information to predict part weights. 
	
	Specifically, the proposed relative attention network consists of two fully connected layers and a Softmax layer. See the bottom part of Figure~\ref{fig:model1}.
	The network takes in 4 pairs of feature vectors, and the Softmax layer output 4 normalized attention weights. 
	To train the attention network, we employ a cosine embedding verification loss. Given an object pair $(i,j)$, the corresponding label $y=1$ if these two samples belong to the same identity, otherwise $y=-1$. Then the loss function is as follows:
	\begin{equation}
		L_{veri} = \left\{
		\begin{array}{ll}
			1 - s(i,j)  &{\rm if} \ y = 1  \\
			max(0, s(i,j) + \alpha)  &{\rm if} \ y = -1   
		\end{array},
		\right.
	\end{equation}
	where $\alpha$ is the margin term. This loss term builds a margin between positive and negative pairs, and thus safeguards the discrminativeness of the embedded features. During training, the cosine embedding loss is jointly optimized with OIM loss.
	For all the context pairs, we select the top $K$ matched pairs as positive contexts, which are utilized for further feature learning. 

	\subsection{Contextual Graph Representation Learning}
	In this section, we introduce the detailed structure of the proposed context graph as well as the GCN model employed to learn graph parameters. The overall structure is illustrated on the right side of Figure~\ref{fig:model2}.
	Given two images $A$ and $B$. The objective of our model is to judge whether target pair appeared in the images ${A_0}$ and ${B_0}$ (in red bounding boxes) belong to the same identity, given $K$ context pairs $({A_i},{B_i}), i\in \{1,..,K\}$. The objective is to construct a graph to jointly take the target pairs and the context information into consideration, and eventually outputs the similarity score. 
	A straightforward solution is to employ two graphs to model each image, and to utilize a Siamese GCN structure to extract features of both graphs, as in~\cite{DBLP:conf/miccai/KtenaPFRLGR17}. However, the Siamese structure prevents the contextual information to propagate between graphs, which leads to significant information loss. 
	In our situation, the targets and contexts all appear in pairs. Therefore, we build a graph whose nodes consist of instance pairs. In this graph, the target node is the center of the graph, which is connected to all the context nodes for information propagation and feature updation. 
	
	In particular, considering a graph $\mathcal{G}=\{\mathcal{V},\mathcal{E}\}$ consisting of $N$ vertices $\mathcal{V}$ and a set of edges $\mathcal{E}$. We assign each node with a pair of features ($\mathbf{x}_{A_j},\mathbf{x}_{B_j}$), $j\in\{0,...,K\}$. If the images have $K$ context pairs, then $N = K+1$. We use $\mathbf{X}\in \mathbb{R}^{N\times 2d}$, where $d$ is the instance-level feature dimension. We use $\mathbf{A} \in \mathbb{R}^{N\times N}$ to denote the adjacent matrix associated with graph $\mathcal{G}$. If we assign the target node as the first node in the graph, then the adjacent matrix is:
	\begin{equation}
		A_{i,j} = \left\{
		\begin{array}{ll}
			1  &{\rm if} \ i = 1 \ {\rm or} \ j = 1  \ {\rm or} \ i = j\\
			0  & \ otherwise   
		\end{array},
		\right.
	\end{equation}
	where $i,j \in \{1,...,N\}$.
	If we use $\mathbf{\hat{A}}$ to denote the normalized adjacency matrix, layer-wise GCN propagates as follows:
	\begin{equation}
		\mathbf{Z}^{(l+1)} = \sigma(\mathbf{\hat{A}Z}^{(l)}\mathbf{W}^{(l)}),
	\end{equation}
	where $\mathbf{Z}^{(l)}$ is the matrix activation of the $l$-th layer, and $\mathbf{Z}^{(0)}=\mathbf{X}$ as input. $\mathbf{W}^{(l)}$ is the learnable parameter matrix and $\sigma$ is the ReLU activation function in our framework. 
	Finally, we utilize a fully connected layer to merge all the vertices features into 1024-dimensional feature vector. And a binary Softmax loss layer is employed supervise network training.
	
	\subsection{Implementation Details}
	To train the detection and instance feature learning network, we use an ImageNet~\cite{DBLP:conf/cvpr/DengDSLL009} pretrained ResNet-50 model to initialize model weights. All the training images are resized to $720 \times 576$, and we use a batch size of $32$. For the graph learning model, we set the layer number of GCN to $3$. The initial learning rate is $0.1$ and is reduced by a factor of 2 after $10$ epochs, and the total training epoch is $20$. We implement our model on Pytorch, and all the models are trained and tested on a TITAN X GPU. For both training and testing, when the scene contains less than $K$ context instances, we randomly replicate the context persons to build $K$ context pairs. For the extreme case where the scene only contains a single instance(\ie, no context exists), these images are not utilized to train the graph model.

	\section{Experiments}
	In this section, we conduct experiments on two widely utilized person search datasets. 
	We first give the experiment settings on different datasets. Then we analyze the contributions of different framework components. We further compare our model with previous state-of-the arts.
	
	\subsection{Dataset and Experiment Setup}
	We evaluate the proposed method on both CUHK-SYSU~\cite{DBLP:conf/cvpr/XiaoLWLW17} and PRW dataset~\cite{DBLP:conf/cvpr/ZhengZSCYT17}.
	CUHK-SYSU dataset is a large-scale person search dataset which contains totally 18184 images, with 8432 different persons and 96143 annotated bounding boxes indicating the locations of different pedestrians. This dataset covers diverse scenes, where 12490 images are collected from real street snap and the rest images comes from movies or TVs. For each query person, there exists at least two images containing the target person in gallery set. In addition, this dataset includes large variations in lighting, occlusion, background and resolution, which are close to the real application scenarios. The whole dataset are officially split into training set and testing set. The training set contains 5532 persons in 11206 images, and the testing set contains 2900 query persons and 6978 images. 
	The PRW dataset is captured on a university campus. It consists of 11816 video frames from 6 cameras, and all these images are manually annotated, resulting in 43110 pedestrian bounding boxes. The number of person IDs is 932. The training set contains 5134 images and 482 different persons, and the testing set contains 6112 images and 2057 query persons. 
	
	We utilize the same evaluation protocol as in ~\cite{DBLP:conf/cvpr/XiaoLWLW17}, where the mean Average Precision (mAP) and the top-1 matching rate are employed as evaluation metrics. The top-1 matching rate is similar with that in person re-identification. The main difference is that a matching is accepted only if the overlap between the bounding box of ground truth person and top-1 matching box is larger than 0.5. 
	
	\begin{table}[]\centering
		\caption{Experiment results using different body parts.}
		\label{AS_part}
		\begin{tabular}{c|cc|cc}
			\hline
			\textbf{Dataset} & \multicolumn{2}{c|}{\textbf{CUHK-SYSU}} & \multicolumn{2}{c}{\textbf{PRW}} \\ \hline
			\textbf{Regions} &{\textbf{top-1($\mathbf{\%}$)}}       & {\textbf{mAP($\mathbf{\%}$)}}        & {\textbf{top-1($\mathbf{\%}$)}}        &{\textbf{mAP($\mathbf{\%}$)}}       \\ \hline
			upper   & 65.7         & 58.3       & 57.6            & 20.3         \\ 
			middle   & 64.3         & 56.1       & 56.8            & 20.2         \\ 
			lower   & 61.8         & 57.0       & 55.9            & 19.6         \\ 
			whole   & 77.5         & 71.8       & 62.8            & 23.9         \\ \hline
			uniform & 79.2         & 75.9       & 65.4            & 24.5         \\ \hline
		\end{tabular}
	\end{table}
	
	\subsection{Effective of Part-based Learning Framework}
	Part-based learning framework has proven to be effective for many person re-id models. To validate the effectiveness of part-based model in our person search framework, we analyze the performance of our model utilizing different part features. The results are reported in Table~\ref{AS_part}. We split the whole body into three parts, with respect to the upper, middle and lower regions of human body, and we utilize our part-based pooling layers to obtain the final part features. We also pool all features in the whole detection box as global feature. We first directly utilize these learned features of different parts to perform the person matching and the results are shown in the first 4 rows in Table~\ref{AS_part}. On CUHK-SYSU dataset, we observe that the performance using middle part feature is comparable with that of upper part feature ($64.3\%$ \vs $65.7\%$), but the lower part features produce a much lower result ($61.8\%$) than the other two parts. These results suggest that the upper and middle parts of a person contain more discriminative information than lower regions.
	We also observe that using global information achieves better results than either parts, which suggests that each part contains unique discriminative information.
	In addition, when we uniformly combine these four features (last row in Table~\ref{AS_part}), \ie, setting $w_r$ in Equation~\ref{equ_relative} to $0.25$ for all parts, the performance is further improved $1.7\%$ (from $77.5\%$ to $79.2\%$). The results on PRW dataset is consistent with that on CUHK-SYSU. These results suggest that region-based features contains discriminative information and can obtain better performance with appropriate aggregation methods.  
	
	\begin{table}[]\centering
		\caption{Component analysis results.}
		\label{AS_context}
		\begin{tabular}{c|cc|cc}
			\hline
			\textbf{Dataset} & \multicolumn{2}{c|}{\textbf{CUHK-SYSU}} & \multicolumn{2}{c}{\textbf{PRW}} \\ \hline
			\textbf{Methods} &{\textbf{top-1($\mathbf{\%}$)}}       & {\textbf{mAP($\mathbf{\%}$)}}        & {\textbf{top-1($\mathbf{\%}$)}}        &{\textbf{mAP($\mathbf{\%}$)}}       \\ \hline
			uniform & 79.2         & 75.9        & 65.4           & 24.5         \\ 
			attention&82.7         & 80.2       & 67.8            & 27.8         \\ 
			graph  &86.5        & 84.1       & 73.6            & 33.4         \\ \hline
		\end{tabular}
	\end{table}
	
	\subsection{Instance Expansion and Contextual Learning}
	
\begin{figure}[t]
		\centering
		\includegraphics[width=\linewidth]{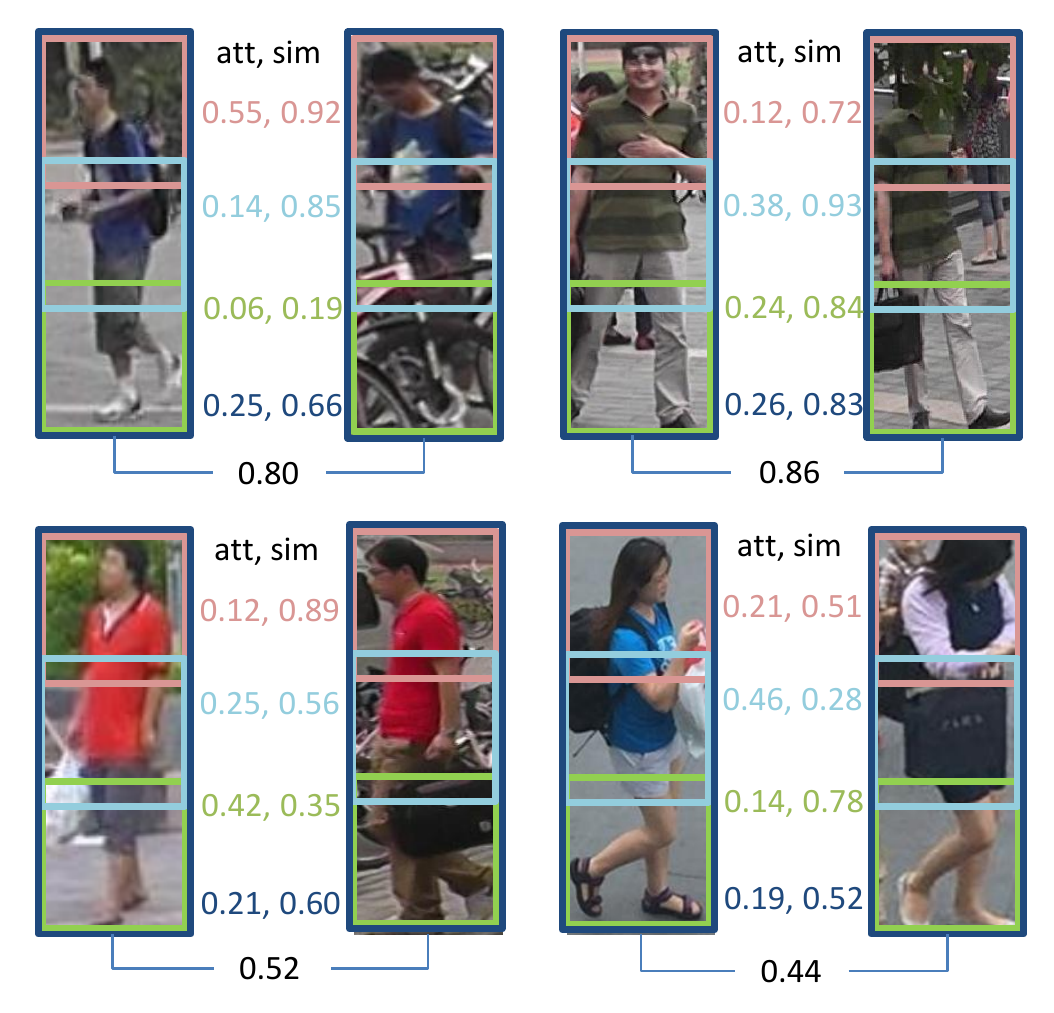}
		\caption{Attention examples. ``att" denotes the attention wights of different body parts, ``sim" denotes similarity between body parts. The values under image pairs denote the overall similarity.}
		\label{fig:att_exp}
	\end{figure}

	With the learned part-based features, we further analyze the contribution of the contextual instance expansion and contextual graph representation learning components. Instead of assigning uniform weights to combine the learned features, we train our relative attention module and assign the learned weight for pairwise parts to measure the overall similarity. In other words, the weights $w_{r}$ in Equation~\ref{equ_relative} are learned based on the feature pairs. Using the learned metric, the retrieval results are shown in Table~\ref{AS_context}. Compared with uniform weights, our relative attention module achieves about 2\% improvement on rank-1 matching rate on both datasets ($79.2\%$ $\rightarrow$ $82.7\%$, $65.4\%$ $\rightarrow$ $67.8\%$,). These results validate the effectiveness of the proposed relative attention module. 
	We also provide attention results in Figure~\ref{fig:att_exp}. The first row shows two positive pairs with partial occlusion. We observe that the bottom part and the top part of the second persons are occluded, respectively. Accordingly, we also observe that the attention weights of the occluded regions (0.06, 0.12) are much lower than other regions. Hence, although the occluded parts have low similarity scores, it does not affect the overall similarity of the target pairs.
	The second row shows two negative pairs. The upper parts in the first example and the lower parts in the second example displays high appearance similarity, which are adaptively assigned with low attention by the relative attention model, and thus results in low overall similarity. 
	
\begin{figure}[t]
		\centering
		\includegraphics[width=0.49\linewidth,height=0.4\linewidth]{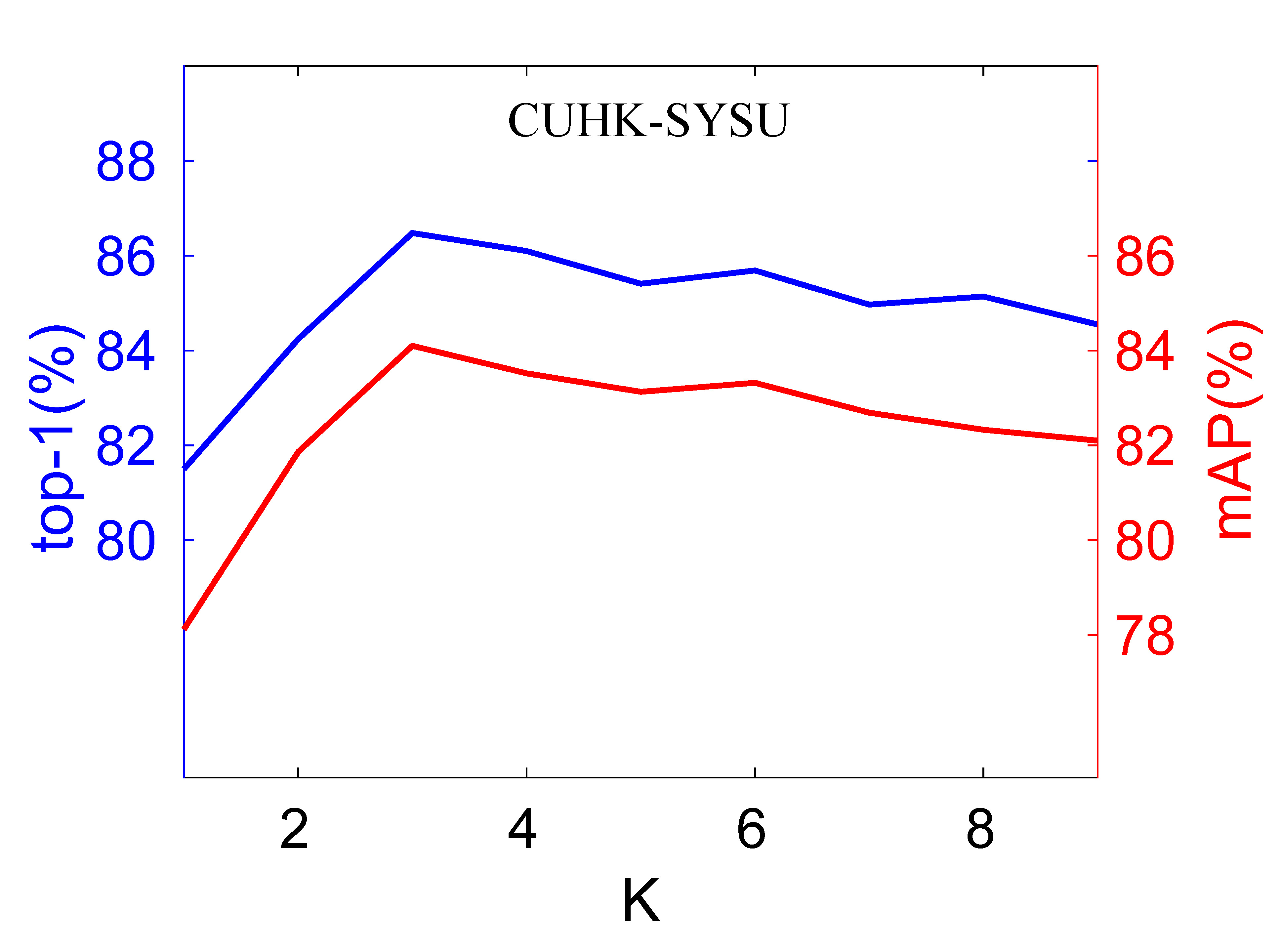}
		\includegraphics[width=0.49\linewidth,height=0.4\linewidth]{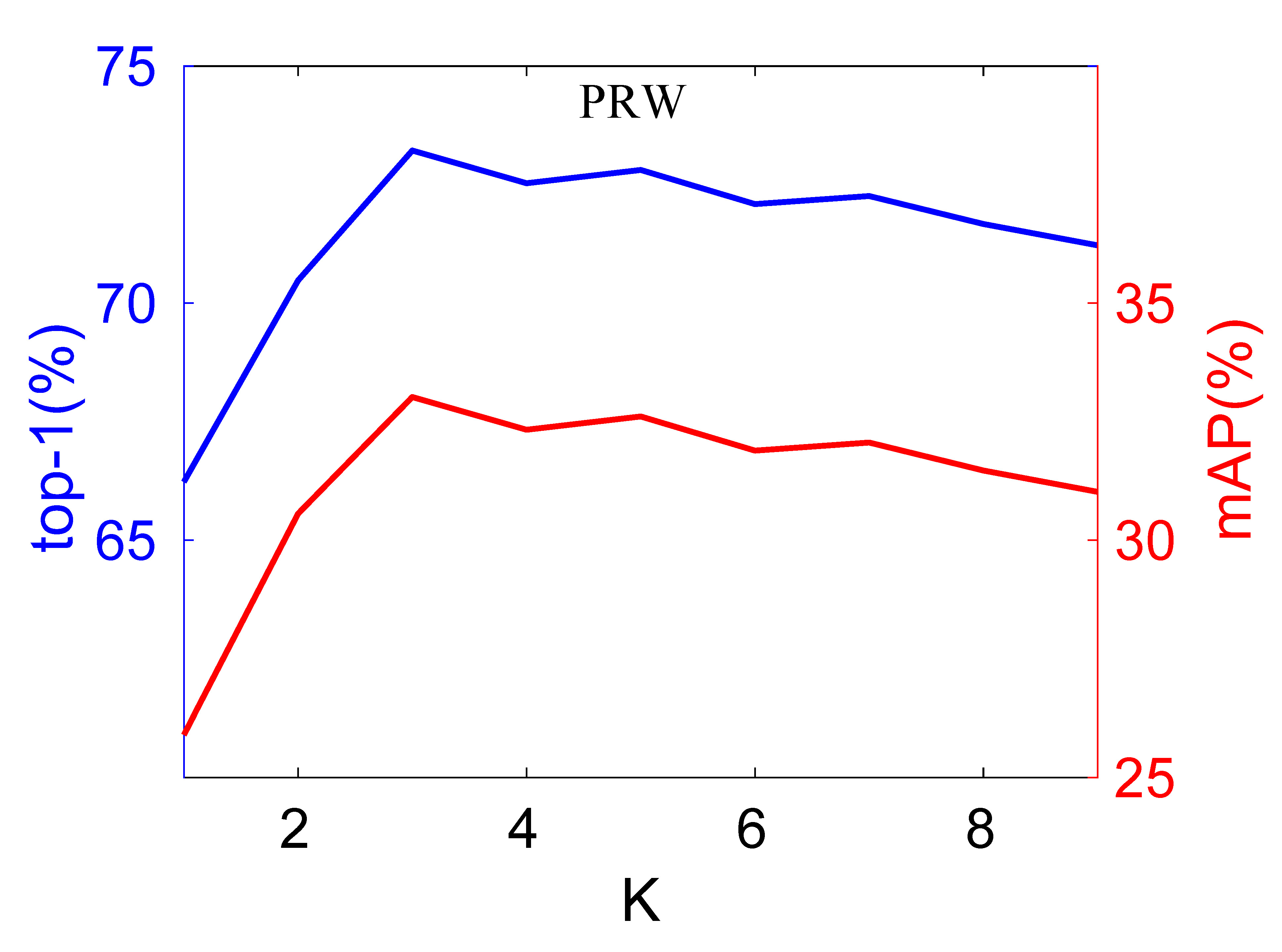}
		\caption{The impact of context size $K$ on performance}
		\label{fig:c_size}
	\end{figure}

	We expand individual features with the top-K matched context pairs, and all these features are modeled by a contextual graph for global representation learning. The results are shown in the last row of Table~\ref{AS_context}. With $K=3$, the contextual graph achieves 86.5\% rank-1 accuracy and 84.1\% mAP on CUHK-SYSU, and 73.6\% rank-1 accuracy and 33.4\% mAP on PRW. These results further demonstrate that by utilizing a graph model to aggregate context information, the proposed model can produce more discriminative representations for person search task. In addition, we also experiment our module with different number of context neighbors. The results are visualized in Figure~\ref{fig:c_size}. From the results we find that the performance will first increase as the number of context neighbors grows. This is because that the first few context pairs are highly confident ones, and thus brings useful information to identify the target pair. After the model gets the best performance with 3 context neighbors, the performance will drop even with more context neighbors, which is due to the existence of more noise context pairs. The comparative results suggest that with appropriate number of context neighbors, our graph model can effectively learn from contextual information and make more accurate judgments.
	
	\subsection{Comparison with State-of-the-Art Methods}
	In this subsection, we report the person search results of our model on both CUHK-SYSU and PRW datasets and compare our model with several state-of-the-art methods, such as IAN~\cite{DBLP:journals/corr/XiaoXTHWF17}, OIM~\cite{DBLP:conf/cvpr/XiaoLWLW17}, I-Net~\cite{DBLP:journals/corr/abs-1804-00376}, NPSM~\cite{DBLP:conf/iccv/LiuFJKZQJY17} and MGTS\cite{DBLP:conf/eccv/ChenZOYT18}. We also compare the results using some other hand-crafted re-id features as reported in~\cite{DBLP:journals/corr/abs-1804-00376}.
	
	\textbf{Results on CUHK dataset.} The comparative results are reported in Table~\ref{Comparison_CUHK_SYSU}. The gallery size of all methods are set to 100. We use ``CNN" to denote the Faster R-CNN detector based on ResNet-50, and we use ``CNN$_{v}$" to denote a VGG-based detector. 
	Compared with results using hand-crafted features, we observe that all the deep learning based feature learning methods achieves significant improvements, which verifies the superiority of deep CNN on person re-id task.
	The original OIM~\cite{DBLP:conf/cvpr/XiaoLWLW17} can be viewed as an overall baseline of the proposed framework, our framework achieves about 8\% improvements on both mAP and top-1 matching rate, which demonstrates effectiveness of the proposed framework.
	IAN~\cite{DBLP:journals/corr/XiaoXTHWF17} proposes a center loss to improve intra-class feature compactness. However, the improvement over OIM is limited on ResNet-50, more improvements can be achieved using stronger backbone networks (ResNet-101). In contrast, our framework only employs ResNet-50 as backbone network, but we still achieve more than 6\% improvements over IAN with stronger backbone. 
	NPSM~\cite{DBLP:conf/iccv/LiuFJKZQJY17} also considers context cues in the gallery images, it proposes a recursive attention framework to sequentially search for the target in the gallery images. Our framework considers context cues in both query and gallery images, thus our method achieves more significant improvements.
	I-Net~\cite{DBLP:journals/corr/abs-1804-00376} introduces a Siamese structure and a novel online pairing loss to learn robust feature representation, I-Net achieves 4\% top-1 accuracy over OIM with hard negative mining. With contextual learning, our method does not need explicit loss designing and negative mining, while we still achieves improvement over I-Net.
	MGTS\cite{DBLP:conf/eccv/ChenZOYT18} designs a segmentation network to generate clean foreground objects, and it utilizes both detection and segmentation objects for two-stream feature learning. This method achieves great performance on CUHK-SYSU dataset, while our method achieves slightly better performance by employing context features.
	
\begin{figure}[t]
		\centering
		\includegraphics[width=0.9\linewidth]{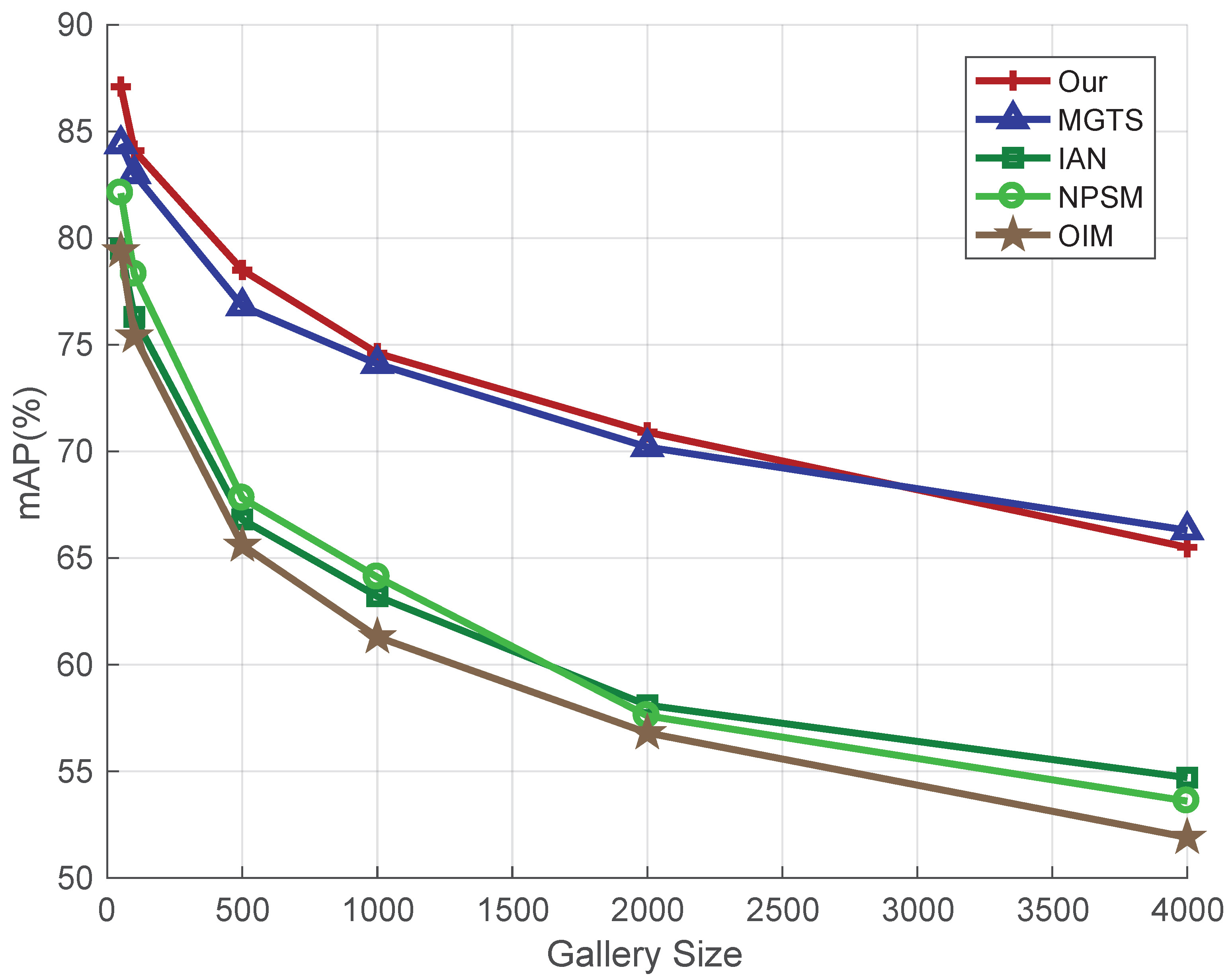}
		\caption{The impact of gallery size on performance}
		\label{fig:g_size}
	\end{figure}

	Instead of fixing the gallery size to 100, we also evaluate our model with different gallery size of [50, 100, 500, 1000, 2000, 4000] and compare with other methods. The results are visualized in Figure~\ref{fig:g_size}. We observe that the performance of all methods will degenerate as the gallery size grows. However, our method still outperforms other methods under different gallery size, which demonstrates the robustness of our proposed method.
	
		\begin{table}[t]
		\centering
		\caption{Comparison of results on CUHK-SYSU with gallery size of 100}
		\begin{tabular}{l|cc}
			\hline
			\textbf{Method} & \multicolumn{1}{c}{\textbf{mAP($\mathbf{\%}$)}} & \multicolumn{1}{c}{\textbf{top-1($\mathbf{\%}$)}} \\
			\hline
			CNN + DSIFT + Euclidean \cite{DBLP:conf/cvpr/ZhaoOW13}& 34.5  & 39.4 \\
			CNN + DSIFT + KISSME \cite{DBLP:conf/cvpr/ZhaoOW13}\cite{DBLP:conf/cvpr/KostingerHWRB12}& 47.8  & 53.6  \\
			CNN + BoW + Cosine \cite{DBLP:conf/iccv/ZhengSTWWT15}& 56.9  & 62.3  \\
			CNN + LOMO + XQDA\cite{DBLP:conf/cvpr/LiaoHZL15} & 68.9  & 74.1  \\
			\hline
			OIM~\cite{DBLP:conf/cvpr/XiaoLWLW17}   & 75.5  & 78.7  \\
			IAN (ResNet-50)~\cite{DBLP:journals/corr/XiaoXTHWF17}   & 76.3  & 80.1  \\
			IAN (ResNet-101)~\cite{DBLP:journals/corr/XiaoXTHWF17}   & 77.2  & 80.5  \\
			NPSM~\cite{DBLP:conf/iccv/LiuFJKZQJY17}  & 77.9  & 81.2  \\
			I-Net~\cite{DBLP:journals/corr/abs-1804-00376} & 79.5 &81.5  \\
			{CNN$_v$ + MGTS}\cite{DBLP:conf/eccv/ChenZOYT18} & {83.0}  & {83.7}  \\
			\hline
			Ours & \textbf{84.1} & \textbf{86.5} \\
			
			\hline
		\end{tabular}%
		\label{Comparison_CUHK_SYSU}%
	\end{table}%
	
	\begin{table}[]
		\centering
		\caption{Comparison of results on PRW}
		\begin{tabular}{l|cc}
			\hline
			\textbf{Method} & \multicolumn{1}{c}{\textbf{mAP($\mathbf{\%}$)}} & \multicolumn{1}{c}{\textbf{top-1($\mathbf{\%}$)}} \\
			\hline
			OIM~\cite{DBLP:conf/cvpr/XiaoLWLW17}   & 21.3  & 49.9  \\
			IAN (ResNet-101)~\cite{DBLP:journals/corr/XiaoXTHWF17} & 23.0  & 61.9  \\
			NPSM~\cite{DBLP:conf/iccv/LiuFJKZQJY17}  & 24.2  & 53.1  \\
			{CNN$_v$ + MGTS}\cite{DBLP:conf/eccv/ChenZOYT18}  & {32.6}  & {72.1} \\
			
			
			\hline
			Ours & \textbf{33.4} & \textbf{73.6} \\
			
			\hline
		\end{tabular}%
		\label{Comparison_PRW}%
	\end{table}%

	\textbf{Results on PRW dataset.} We compare our model with stat-of-the-art methods on PRW dataset, the results are shown in Table~\ref{Comparison_PRW}. Compared with CUHK-SYSU dataset, all the methods achieves poorer results on PRW dataset, especially on mAP. This is mainly due to less training data on PRW dataset which limits the generalization capability of the learned model. On this dataset, our model achieves 73.6\% top-1 accuracy and 33.4 mAP, which outperforms previous state-of-the-arts.

	\subsection{Discussion on Graph Structures}
	Siamese structures~\cite{DBLP:conf/cvpr/LiZXW14,DBLP:conf/cvpr/AhmedJM15} have been widely adopted in person re-id task to estimate the similarity between target pairs. Recently, some works~\cite{DBLP:conf/miccai/KtenaPFRLGR17,DBLP:journals/corr/abs-1802-07459} have extended the Siamese structure to GCN for measuring the distance between graphs. Following this idea, we build two separate graphs for each image pair. Then we utilize two shared weights GCN to learn each graph, and the learned features are combined to output the similarity between the two graphs. The overall idea of such graph is given in the bottom part of Figure~\ref{fig:intro} The results are reported in Table~\ref{tab:gs}. We observe that the Siamese graph learning structure also achieves improvements over baseline models, but our graph achieves better performance. This is due to our design that the pairwise relations have been contained within each graph node, and thus making the graph easier to capture context information. These results also demonstrate that graph structures play an important role in learning context features, and it is a promising future direction to further improve the performance of person search framework.

	\begin{table}[]\centering
		\caption{Results of different graph structures}
		\label{tab:gs}
		\begin{tabular}{c|cc|cc}
			\hline
			\textbf{Dataset} & \multicolumn{2}{c|}{\textbf{CUHK-SYSU}} & \multicolumn{2}{c}{\textbf{PRW}} \\ \hline
			\textbf{Methods} &{\textbf{top-1}}       & {\textbf{mAP}}        & {\textbf{top-1}}        &{\textbf{mAP}}       \\ \hline
			DisGCN~\cite{DBLP:conf/miccai/KtenaPFRLGR17} & 83.4        & 81.3      &  69.8         &   29.5        \\ 
			Ours   &86.5        & 84.1       & 73.6            & 33.4         \\ \hline
		\end{tabular}
	\end{table}
	
	\section{Conclusion}
	In this work, we propose to employ contextual information to improve the robustness of person search results. Experimental results demonstrate that the proposed instance expansion method can effectively find useful contextual information, while the introduced graph learning framework successfully aggregates the contextual information to update the target similarity. The proposed framework achieves state-of-the-arts performance on two widely adopted person search benchmarks.
	
	\section*{Acknowledgments}
	This work was supported by National Science Foundation of China (U1611461,61521062). This work was partly supported by National Key Research and Development Program of China (2016YFB1001003), STCSM(18DZ1112300,18DZ2270700). This work was also partially supported by joint research grant of SJTU-BIGO LIVE, and joint research grant of SJTU-Minivision, and China's Thousand Talent Program.
	
	{\small
		\bibliographystyle{ieee_fullname}
		\bibliography{egbib}

\begin{thebibliography}{10}\itemsep=-1pt

\bibitem{DBLP:conf/cvpr/AhmedJM15}
Ejaz Ahmed, Michael~J. Jones, and Tim~K. Marks.
\newblock An improved deep learning architecture for person re-identification.
\newblock In {\em CVPR}, pages 3908--3916, 2015.

\bibitem{DBLP:conf/eccv/AssariIS16}
Shayan~Modiri Assari, Haroon Idrees, and Mubarak Shah.
\newblock Human re-identification in crowd videos using personal, social and
  environmental constraints.
\newblock In {\em ECCV}, pages 119--136, 2016.

\bibitem{BEDAGKARGALA20121908}
A. Bedagkar-Gala and Shishir~K. Shah.
\newblock Part-based spatio-temporal model for multi-person re-identification.
\newblock {\em Pattern Recognition Letters}, 33(14):1908 -- 1915, 2012.

\bibitem{DBLP:conf/iccvw/CaoCHP17}
Min Cao, Chen Chen, Xiyuan Hu, and Silong Peng.
\newblock From groups to co-traveler sets: Pair matching based person
  re-identification framework.
\newblock In {\em ICCV Workshops}, pages 2573--2582, 2017.

\bibitem{DBLP:conf/cvpr/Chen0LSW18}
Dapeng Chen, Dan Xu, Hongsheng Li, Nicu Sebe, and Xiaogang Wang.
\newblock Group consistent similarity learning via deep {CRF} for person
  re-identification.
\newblock In {\em CVPR}, pages 8649--8658, 2018.

\bibitem{DBLP:conf/cvpr/ChenYCZ16}
Dapeng Chen, Zejian Yuan, Badong Chen, and Nanning Zheng.
\newblock Similarity learning with spatial constraints for person
  re-identification.
\newblock In {\em CVPR}, pages 1268--1277, 2016.

\bibitem{DBLP:conf/eccv/ChenZOYT18}
Di Chen, Shanshan Zhang, Wanli Ouyang, Jian Yang, and Ying Tai.
\newblock Person search via a mask-guided two-stream {CNN} model.
\newblock In {\em ECCV}, pages 764--781, 2018.

\bibitem{DBLP:conf/cvpr/ChenCZH17}
Weihua Chen, Xiaotang Chen, Jianguo Zhang, and Kaiqi Huang.
\newblock Beyond triplet loss: {A} deep quadruplet network for person
  re-identification.
\newblock In {\em CVPR}, pages 1320--1329, 2017.

\bibitem{CHENG2018}
De Cheng, Yihong Gong, Zhihui Li, Dingwen Zhang, Weiwei Shi, and Xingjun Zhang.
\newblock Cross-scenario transfer metric learning for person re-identification.
\newblock {\em Pattern Recognition Letters}, 2018.

\bibitem{DBLP:conf/cvpr/DengDSLL009}
Jia Deng, Wei Dong, Richard Socher, Li{-}Jia Li, Kai Li, and Fei{-}Fei Li.
\newblock Imagenet: {A} large-scale hierarchical image database.
\newblock In {\em CVPR}, pages 248--255, 2009.

\bibitem{DBLP:conf/cvpr/FarenzenaBPMC10}
Michela Farenzena, Loris Bazzani, Alessandro Perina, Vittorio Murino, and Marco
  Cristani.
\newblock Person re-identification by symmetry-driven accumulation of local
  features.
\newblock In {\em CVPR}, pages 2360--2367, 2010.

\bibitem{DBLP:journals/corr/abs-1711-04043}
Victor Garcia and Joan Bruna.
\newblock Few-shot learning with graph neural networks.
\newblock In {\em ICLR}, 2018.

\bibitem{DBLP:conf/eccv/GrayT08}
Douglas Gray and Hai Tao.
\newblock Viewpoint invariant pedestrian recognition with an ensemble of
  localized features.
\newblock In {\em ECCV}, pages 262--275, 2008.

\bibitem{DBLP:conf/cvpr/HeZRS16}
Kaiming He, Xiangyu Zhang, Shaoqing Ren, and Jian Sun.
\newblock Deep residual learning for image recognition.
\newblock In {\em CVPR}, pages 770--778, 2016.

\bibitem{DBLP:journals/corr/abs-1804-00376}
Zhenwei He, Lei Zhang, and Wei Jia.
\newblock End-to-end detection and re-identification integrated net for person
  search.
\newblock {\em CoRR}, abs/1804.00376, 2018.

\bibitem{DBLP:journals/corr/abs-1806-03084}
Qingqiu Huang, Yu Xiong, and Dahua Lin.
\newblock Unifying identification and context learning for person recognition.
\newblock In {\em CVPR}, pages 2217--2225, 2018.

\bibitem{DBLP:journals/corr/KipfW16}
Thomas~N. Kipf and Max Welling.
\newblock Semi-supervised classification with graph convolutional networks.
\newblock In {\em ICLR}, 2017.

\bibitem{DBLP:conf/cvpr/KostingerHWRB12}
Martin K{\"{o}}stinger, Martin Hirzer, Paul Wohlhart, Peter~M. Roth, and Horst
  Bischof.
\newblock Large scale metric learning from equivalence constraints.
\newblock In {\em CVPR}, pages 2288--2295, 2012.

\bibitem{DBLP:conf/nips/KrizhevskySH12}
Alex Krizhevsky, Ilya Sutskever, and Geoffrey~E. Hinton.
\newblock Imagenet classification with deep convolutional neural networks.
\newblock In {\em NIPS}, pages 1106--1114, 2012.

\bibitem{DBLP:conf/miccai/KtenaPFRLGR17}
Sofia~Ira Ktena, Sarah Parisot, Enzo Ferrante, Martin Rajchl, Matthew C.~H.
  Lee, Ben Glocker, and Daniel Rueckert.
\newblock Distance metric learning using graph convolutional networks:
  Application to functional brain networks.
\newblock In {\em MICCAI}, pages 469--477, 2017.

\bibitem{DBLP:conf/cvpr/LiZXW14}
Wei Li, Rui Zhao, Tong Xiao, and Xiaogang Wang.
\newblock Deepreid: Deep filter pairing neural network for person
  re-identification.
\newblock In {\em CVPR}, pages 152--159, 2014.

\bibitem{DBLP:conf/cvpr/LiaoHZL15}
Shengcai Liao, Yang Hu, Xiangyu Zhu, and Stan~Z. Li.
\newblock Person re-identification by local maximal occurrence representation
  and metric learning.
\newblock In {\em CVPR}, pages 2197--2206, 2015.

\bibitem{DBLP:journals/tip/LinSYXWWL17}
Weiyao Lin, Yang Shen, Junchi Yan, Mingliang Xu, Jianxin Wu, Jingdong Wang, and
  Ke Lu.
\newblock Learning correspondence structures for person re-identification.
\newblock {\em {IEEE} Trans. Image Processing}, 26(5):2438--2453, 2017.

\bibitem{DBLP:conf/iccv/LisantiMBF17}
Giuseppe Lisanti, Niki Martinel, Alberto~Del Bimbo, and Gian~Luca Foresti.
\newblock Group re-identification via unsupervised transfer of sparse features
  encoding.
\newblock In {\em ICCV}, pages 2468--2477, 2017.

\bibitem{DBLP:journals/corr/abs-1802-07459}
Bang Liu, Ting Zhang, Di Niu, Jinghong Lin, Kunfeng Lai, and Yu Xu.
\newblock Matching long text documents via graph convolutional networks.
\newblock {\em CoRR}, abs/1802.07459, 2018.

\bibitem{DBLP:conf/iccv/LiuFJKZQJY17}
Hao Liu, Jiashi Feng, Zequn Jie, Jayashree Karlekar, Bo Zhao, Meibin Qi,
  Jianguo Jiang, and Shuicheng Yan.
\newblock Neural person search machines.
\newblock In {\em ICCV}, pages 493--501, 2017.

\bibitem{DBLP:conf/cvpr/LiuNYZCH18}
Jinxian Liu, Bingbing Ni, Yichao Yan, Peng Zhou, Shuo Cheng, and Jianguo Hu.
\newblock Pose transferrable person re-identification.
\newblock In {\em CVPR}, pages 4099--4108, 2018.

\bibitem{DBLP:journals/ijcv/Lowe04}
David~G. Lowe.
\newblock Distinctive image features from scale-invariant keypoints.
\newblock {\em International Journal of Computer Vision}, 60(2):91--110, 2004.

\bibitem{DBLP:conf/eccv/LuMNYRY18}
Xiankai Lu, Chao Ma, Bingbing Ni, Xiaokang Yang, Ian~D. Reid, and Ming{-}Hsuan
  Yang.
\newblock Deep regression tracking with shrinkage loss.
\newblock In {\em ECCV}, pages 369--386, 2018.

\bibitem{DBLP:conf/avss/MazzonPC13}
Riccardo Mazzon, Fabio Poiesi, and Andrea Cavallaro.
\newblock Detection and tracking of groups in crowd.
\newblock In {\em AVSS}, pages 202--207, 2013.

\bibitem{DBLP:conf/eccv/QiWJSZ18}
Siyuan Qi, Wenguan Wang, Baoxiong Jia, Jianbing Shen, and Song{-}Chun Zhu.
\newblock Learning human-object interactions by graph parsing neural networks.
\newblock In {\em ECCV}, pages 407--423, 2018.

\bibitem{DBLP:conf/nips/RenHGS15}
Shaoqing Ren, Kaiming He, Ross~B. Girshick, and Jian Sun.
\newblock Faster {R-CNN:} towards real-time object detection with region
  proposal networks.
\newblock In {\em NIPS}, pages 91--99, 2015.

\bibitem{DBLP:conf/cvpr/SchroffKP15}
Florian Schroff, Dmitry Kalenichenko, and James Philbin.
\newblock Facenet: {A} unified embedding for face recognition and clustering.
\newblock In {\em CVPR}, pages 815--823, 2015.

\bibitem{DBLP:conf/eccv/ShenLYCW18}
Yantao Shen, Hongsheng Li, Shuai Yi, Dapeng Chen, and Xiaogang Wang.
\newblock Person re-identification with deep similarity-guided graph neural
  network.
\newblock In {\em ECCV}, pages 508--526, 2018.

\bibitem{DBLP:conf/iccv/ShenLYXWW15}
Yang Shen, Weiyao Lin, Junchi Yan, Mingliang Xu, Jianxin Wu, and Jingdong Wang.
\newblock Person re-identification with correspondence structure learning.
\newblock In {\em ICCV}, pages 3200--3208, 2015.

\bibitem{DBLP:conf/mm/SongNYRXY17}
Zhichao Song, Bingbing Ni, Yichao Yan, Zhe Ren, Yi Xu, and Xiaokang Yang.
\newblock Deep cross-modality alignment for multi-shot person
  re-identification.
\newblock In {\em ACM MM}, pages 645--653, 2017.

\bibitem{DBLP:conf/iccv/SuLZX0T17}
Chi Su, Jianing Li, Shiliang Zhang, Junliang Xing, Wen Gao, and Qi Tian.
\newblock Pose-driven deep convolutional model for person re-identification.
\newblock In {\em ICCV}, pages 3980--3989, 2017.

\bibitem{DBLP:journals/tip/WangSS18}
Wenguan Wang, Jianbing Shen, and Ling Shao.
\newblock Video salient object detection via fully convolutional networks.
\newblock {\em {IEEE} Trans. Image Processing}, 27(1):38--49, 2018.

\bibitem{DBLP:conf/eccv/WangG18}
Xiaolong Wang and Abhinav Gupta.
\newblock Videos as space-time region graphs.
\newblock In {\em ECCV}, pages 413--431, 2018.

\bibitem{DBLP:conf/mm/WeiZY0T17}
Longhui Wei, Shiliang Zhang, Hantao Yao, Wen Gao, and Qi Tian.
\newblock {GLAD:} global-local-alignment descriptor for pedestrian retrieval.
\newblock In {\em ACM MM}, pages 420--428, 2017.

\bibitem{DBLP:journals/corr/XiaoXTHWF17}
Jimin Xiao, Yanchun Xie, Tammam Tillo, Kaizhu Huang, Yunchao Wei, and Jiashi
  Feng.
\newblock {IAN:} the individual aggregation network for person search.
\newblock {\em Pattern Recognition}, 87:332--340, 2019.

\bibitem{DBLP:conf/cvpr/XiaoLWLW17}
Tong Xiao, Shuang Li, Bochao Wang, Liang Lin, and Xiaogang Wang.
\newblock Joint detection and identification feature learning for person
  search.
\newblock In {\em CVPR}, pages 3376--3385, 2017.

\bibitem{DBLP:conf/mm/XuMHL14}
Yuanlu Xu, Bingpeng Ma, Rui Huang, and Liang Lin.
\newblock Person search in a scene by jointly modeling people commonness and
  person uniqueness.
\newblock In {\em {ACM} {MM}}, pages 937--940, 2014.

\bibitem{yan2018multi}
Yichao Yan, Bingbing Ni, Jinxian Liu, and Xiaokang Yang.
\newblock Multi-level attention model for person re-identification.
\newblock {\em Pattern Recognition Letters}, 2018.

\bibitem{DBLP:conf/eccv/YanNSMYY16}
Yichao Yan, Bingbing Ni, Zhichao Song, Chao Ma, Yan Yan, and Xiaokang Yang.
\newblock Person re-identification via recurrent feature aggregation.
\newblock In {\em ECCV}, pages 701--716, 2016.

\bibitem{DBLP:conf/ijcai/YanNY17}
Yichao Yan, Bingbing Ni, and Xiaokang Yang.
\newblock Predicting human interaction via relative attention model.
\newblock In {\em IJCAI}, pages 3245--3251, 2017.

\bibitem{DBLP:conf/mm/YanXNZY17}
Yichao Yan, Jingwei Xu, Bingbing Ni, Wendong Zhang, and Xiaokang Yang.
\newblock Skeleton-aided articulated motion generation.
\newblock In {\em ACM MM}, pages 199--207, 2017.

\bibitem{DBLP:conf/cvpr/ZhangXG16}
Li Zhang, Tao Xiang, and Shaogang Gong.
\newblock Learning a discriminative null space for person re-identification.
\newblock In {\em CVPR}, pages 1239--1248, 2016.

\bibitem{ZHAO2018}
Cairong Zhao, Kang Chen, Zhihua Wei, Yipeng Chen, Duoqian Miao, and Wei Wang.
\newblock Multilevel triplet deep learning model for person re-identification.
\newblock {\em Pattern Recognition Letters}, 2018.

\bibitem{DBLP:conf/cvpr/ZhaoOW13}
Rui Zhao, Wanli Ouyang, and Xiaogang Wang.
\newblock Unsupervised salience learning for person re-identification.
\newblock In {\em CVPR}, pages 3586--3593, 2013.

\bibitem{DBLP:conf/cvpr/ZhaoOW14}
Rui Zhao, Wanli Ouyang, and Xiaogang Wang.
\newblock Learning mid-level filters for person re-identification.
\newblock In {\em CVPR}, pages 144--151, 2014.

\bibitem{DBLP:journals/corr/ZhengHLY17}
Liang Zheng, Yujia Huang, Huchuan Lu, and Yi Yang.
\newblock Pose invariant embedding for deep person re-identification.
\newblock {\em CoRR}, abs/1701.07732, 2017.

\bibitem{DBLP:conf/iccv/ZhengSTWWT15}
Liang Zheng, Liyue Shen, Lu Tian, Shengjin Wang, Jingdong Wang, and Qi Tian.
\newblock Scalable person re-identification: {A} benchmark.
\newblock In {\em ICCV}, pages 1116--1124, 2015.

\bibitem{DBLP:conf/cvpr/ZhengZSCYT17}
Liang Zheng, Hengheng Zhang, Shaoyan Sun, Manmohan Chandraker, Yi Yang, and Qi
  Tian.
\newblock Person re-identification in the wild.
\newblock In {\em CVPR}, pages 3346--3355, 2017.

\bibitem{DBLP:journals/pami/ZhengGX16}
Wei{-}Shi Zheng, Shaogang Gong, and Tao Xiang.
\newblock Towards open-world person re-identification by one-shot group-based
  verification.
\newblock {\em {IEEE} Trans. Pattern Anal. Mach. Intell.}, 38(3):591--606,
  2016.

\bibitem{DBLP:journals/corr/ZhengZY16}
Zhedong Zheng, Liang Zheng, and Yi Yang.
\newblock A discriminatively learned {CNN} embedding for person
  reidentification.
\newblock {\em {TOMCCAP}}, 14(1):13:1--13:20, 2018.

\bibitem{DBLP:conf/cvpr/ZhouHWWT17}
Zhen Zhou, Yan Huang, Wei Wang, Liang Wang, and Tieniu Tan.
\newblock See the forest for the trees: Joint spatial and temporal recurrent
  neural networks for video-based person re-identification.
\newblock In {\em CVPR}, pages 6776--6785, 2017.

\end{thebibliography}
	}
	
\end{document}